\documentclass[journal,article,submit,pdftex,moreauthors]
{Definitions/mdpi} 
\usepackage{algorithm}
\usepackage{algpseudocode}

\pdfoutput=1 

\firstpage{1} 
\pubvolume{1}
\issuenum{1}
\articlenumber{0}
\pubyear{2025}
\copyrightyear{2025}
\datereceived{ } 
\daterevised{ } 
\dateaccepted{ } 
\datepublished{ } 
\hreflink{https://doi.org/} 

\Title{Graph-Level Label-Only Membership Inference Attack against Graph Neural Networks}

\TitleCitation{Graph-Level Label-Only Membership Inference Attack against Graph Neural Networks}

\Author{Jiazhu Dai * and Yubing Lu}

\AuthorNames{Jiazhu Dai, Yubing Lu}

\isAPAStyle{%
       \AuthorCitation{Dai, J., Lu, Y.}
         }{%
        \isChicagoStyle{%
        \AuthorCitation{Dai, Jiazhu, Yubing Lu.}
        }{
        \AuthorCitation{Dai, J., Lu, Y.}
        }
}

\address{%
School of Computer Engineering and Science, Shanghai University, Shanghai 200444, China}

\corres{Correspondence: daijz@shu.edu.cn}

\abstract{Graph neural networks (GNNs) are widely used for graph-structured data but are vulnerable to membership inference attacks (MIAs) in graph classification tasks, which determine if a graph was part of the training dataset, potentially causing data leakage. Existing MIAs rely on prediction probability vectors, but they become ineffective when only prediction labels are available.
We propose a \textbf{G}raph-level \textbf{L}abel-\textbf{O}nly \textbf{M}embership \textbf{I}nference \textbf{A}ttack (GLO-MIA), which is based on the intuition that the target model's predictions on training data are more stable than those on testing data. GLO-MIA generates a set of perturbed graphs for target graph by adding perturbations to its effective features and queries the target model with the perturbed graphs to get their prediction labels, which are then used to calculate robustness score of the target graph. Finally, by comparing the robustness score with a predefined threshold, the membership of the target graph can be inferred correctly with high probability. Our evaluation on three datasets and four GNN models shows that GLO-MIA achieves an attack accuracy of up to 0.825 , outperforming baseline work by 8.5\% and closely matching the performance of probability-based MIAs, even with only prediction labels. }


\begin{document}

\setcounter{section}{0} 
\section{Introduction}
Graph-structured data widely exists in numerous fields of the real world, including biology \cite{1}, social networks \cite{2} and other fields. However, its non-Euclidean nature makes its analysis more challenging. In recent years, Graph Neural Networks (GNNs) have been widely applied to various tasks related to graph structures, such as node classification \cite{3}, link prediction \cite{4,5}, graph classification \cite{6}, etc., due to their ability to capture hidden patterns in graph-structured data, and have achieved state-of-the-art performance. However, the risk of privacy leakage in GNNs has raised significant concerns. 

Previous studies have shown that GNNs are vulnerable to a privacy attack named membership inference attack (MIA) \cite{7,8,9,10,11} where the adversary can infer whether the given data has been used for the training of the model, leading to serious privacy leakage, especially in sensitive applications. For example, in drug research, the training data includes protein interaction networks and molecular properties, the adversary can leverage the model's prediction probability vectors to infer whether specific proteins or molecules are used in the training set, leaking sensitive molecular data or compromising drug development confidentiality. Therefore, it is important and urgent to study the membership inference attack against GNNs. 

Membership inference attacks exploit the model's memorization of the training data, finding the differences in the model's output between the training data (members) and the testing data (non-members) to infer membership. In graph classification tasks, MIAs aim to infer the membership of the given graph. Existing MIAs usually utilize prediction probability vectors to infer membership. However, in real-world scenarios where GNNs may only provide prediction labels rather than probability vectors, these MIAs become ineffective. Therefore, it is necessary to investigate whether there exists MIA threat against GNNs in graph classification tasks when only predicton labels are available to adversary, which, to the best of our knowledge, remains insufficiently explored.

We propose a \textbf{G}raph-level \textbf{L}abel-\textbf{O}nly \textbf{M}embership \textbf{I}nference \textbf{A}ttack (GLO-MIA) against GNNs, revealing the vulnerability of GNNs to such threats. Specifically, given a target graph, GLO-MIA generates a set of perturbed graphs based on the target graph. Then these perturbed graphs are used to query the target model to obtain their prediction labels. Next, the robustness score of the target graph is calculated based on these predicton labels. Finally, the robustness score of the target graph is compared with a predefined threshold. If it is greater than the threshold, the target graph is considered a member, otherwise, it is considered a non-member. In the label-only scenario, the adversary cannot use sufficient model information for membership inference, but the adversary can utilize a shadow dataset with a distribution similar to the target dataset to train a shadow model that mimics the behavior of the target model. We leverage the shadow model to determine the perturbation magnitude and threshold to distinguish members and non-members. Due to the diverse feature distributions and types of graph data, we propose a general perturbation method that applies small, uniformly sampled numerical values to the graph's effective features. 

We evaluate the performance of GLO-MIA on four GNN models and three datasets. The experimental results show that our attack achieves an accuracy of up to 0.825, surpassing the baseline work by up to 8.5\%. Even only using prediction labels, our attack achieves performance close to that of probability-based MIAs. Furthermore, we analyze the impact of perturbation magnitude and the number of perturbed graphs on the attack performance.

In summary, our main contributions are as follows:

\begin{itemize}
    \item To the best of our knowledge, we propose GLO-MIA, the first label-only membership inference attack against graph neural networks in graph classification tasks, revealing the vulnerability of GNNs to such threats.
    \item We propose a general perturbation method to implement GLO-MIA, which infers membership by leveraging the robustness of graph samples to the perturbations.
    \item We evaluate GLO-MIA using four representative GNN methods on three real-world datasets and analyze the factors that influence attack performance. Experimental results show that GLO-MIA can achieve an attack accuracy of up to 0.825, outperforming baselines by up to 8.5\%. Even in the most restrictive black-box scenario, GLO-MIA achieves performance comparable to probability-based MIAs which utilize prediction probability vectors of model.
\end{itemize}

The rest of this paper is organized as follows. Section \ref{s2} introduces the background knowledge of relevant GNNs. Section \ref{s3} discusses the related work. Section \ref{s4} introduces GLO-MIA in detail. Section \ref{s5} presents the evaluation of GLO-MIA. Finally, Section \ref{s6}  concludes the whole paper.


\section{Background} \label{s2}
In this section, we introduce the basic notations of Graph Neural Networks (GNNs) and describe four representative GNN models.
\subsection{Notations }
We define an undirected, unweighted attributed graph as \( G = (V, E, X) \), where \( V = \{v_1, v_2, \dots, v_n\} \) is the set of all nodes in the graph, \( n = |V| \) represents the number of the nodes, and \( E \) is the set of edges, where \( m = |E| \) represents the number of edges. For any two nodes \( v_i, v_j \in V \), \( e_{ij} = (v_i, v_j) \in E \) indicates that there exists an edge between \( v_i \) and \( v_j \). \( A \) is the adjacency matrix of the graph. If \( e_{ij} \in E \), \( a_{ij} = 1 \), otherwise, \( a_{ij} = 0 \). \( X \in \mathbb{R}^{n \times d} \) represents the feature matrix of the nodes in the graph, where \( d \) is the feature dimensionality.
\subsection{Graph Neural Networks}
Graph Neural Networks (GNNs) are a class of neural networks specifically designed to handle graph-structured data. They take the node feature matrix and the graph topology structure as inputs to learn the representation of a single node or the entire graph. Most of the existing GNNs utilize the message passing mechanism. By iteratively performing the "aggregate-update" operation, node embeddings are generated. After \(l\) aggregation iterations, the node representation captures the information of \(l\) -hop neighbors. The \(l\)-th layer of a GNN can be expressed as:
\begin{linenomath}
\begin{equation}
H^{(l)}_{i} = \text{UPDATE} \left( \text{AGGREGATE} \left( H_{i}^{(l-1)} ,\left\{ H_{j}^{(l-1)} \middle| j \in N(i) \right\} \right) \right)
\end{equation}
\end{linenomath}
where \( H^{(l)}_{i} \) represents the embedding of node \( v_i \) at the \( l \)-th layer, \( N(i) \) denotes the set of neighbors of node \( v_i\), \( \text{AGGREGATE}(\cdot) \) aggregates information from node \( v_i \) and its neighbors, and \( \text{UPDATE}(\cdot) \) applies a nonlinear transformation to the aggregated information. The differences among GNN models primarily lie in the implementation strategies of these two functions. We select four GNNs, including Graph Convolutional Network (GCN) \cite{3}, Graph Attention Network (GAT) \cite{12}, SAmple and aggreGatE (GraphSAGE) \cite{13}, and Graph Isomorphism Network (GIN) \cite{6}.

GCN employs symmetric normalization, and the node embedding is represented as:
\begin{linenomath}
\begin{equation}
H_i^{(l)} = \sigma \left( \widetilde{D}^{-\frac{1}{2}} \widetilde{A} \widetilde{D}^{-\frac{1}{2}} H^{(l-1)} W^{(l-1)} \right)
\end{equation}
\end{linenomath}
where \(\widetilde{A} = A + I\) and \(\widetilde{D}_{ii} = \sum_j \widetilde{A}_{ij}\). The embedding of each node aggregates as a weighted average of itself and its neighboring nodes, \(\sigma\) is the activation function, which generates new embeddings from the aggregated results.  

GAT employs an attention mechanism, and the node embedding is represented as:
\begin{linenomath}
\begin{equation}
H_{i}^{(l)} = \sigma \left( \sum_{j \in N(i) \cup i} \alpha_{ij}^{(l)} W^{(l)} H_{j}^{(l-1)} \right)
\end{equation}
\end{linenomath}

where \(\alpha_{ij}\) represents the attention coefficient between node pairs, indicating the strength of the connection between nodes.  

GraphSAGE characterizes the central node by aggregating information from a fixed number of randomly sampled neighbor nodes. It supports various aggregation methods, including mean aggregation, LSTM-based aggregation, and max-pooling aggregation. In our work, we employ mean aggregation, and the node embedding is represented as:
\begin{linenomath}
\begin{equation}
H_{i}^{(l)} = \sigma \left( W^{(l)} \cdot \text{MEAN} \left( H_{i}^{(l-1)},\left\{ H_{j}^{(l-1)} \middle| j \in N(i) \right\} \right) \right)
\end{equation}
\end{linenomath}
where \(\text{MEAN}(\cdot)\) denotes the mean aggregation function.  

GIN are based on graph isomorphism testing, and their aggregation method preserves the strict distinguishability of neighbor features. The current node embedding is summed with the aggregation results and then passed through a multilayer perceptron (MLP):
\begin{linenomath}
\begin{equation}
H_{i}^{(l)} = \text{MLP}^{(l)} \left( 1 + \epsilon^{(l)} \right) H_{i}^{(l-1)} + \sum_{j \in N(i)} H_{j}^{(l-1)}
\end{equation}
\end{linenomath}
where \(\epsilon\) is a learnable parameter used to adjust the weight of the central node.  

After obtaining the final node embeddings, GNNs can perform tasks like node classification, link prediction, and graph classification. This paper focuses on the graph classification, where GNNs aggregate all node embeddings into a single graph-level embedding \(H_{G}\) using a readout layer, represented as:
\begin{linenomath}
\begin{equation}
H_{G} = \text{READOUT}(\{ H_{i}^{(l)} | i \in G \})
\end{equation}
\end{linenomath}
where \(\text{READOUT}(\cdot)\) represents a readout function.  


\section{Related Work}\label{s3}
In this section, we discuss the related work on membership inference attacks(MIAs).

In membership inference attacks, the adversary aims to determine whether any given data record was used in the training of the target model. These attacks are generally categorized into two types: training-based MIAs and threshold-based MIAs. Training-based MIAs use shadow models and shadow datasets to construct attack features and trains an attack model to infer membership. Threshold-based MIAs compare scores (e.g., maximum confidence, prediction cross-entropy) with a predefined threshold to infer membership of data. In the field of traditional machine learning, research on membership inference attacks has been extensively conducted \cite{14,15,16,17,18,19}.

In recent years, membership inference attacks on GNNs have attracted increasing attention. MIAs on GNNs include node-level \cite{7,11}, link-level \cite{20} and graph-level MIAs \cite{9,10}. Node-level MIAs determine the membership of a node. Link-level MIAs focus on determining whether a link between two nodes exists in the training graph. Graph-level MIAs identify the membership of an entire graph. 

In graph-level MIAs, adversary usually utilizes the difference in the output distribution of the target model between members and non-members. Wu et al. \cite{9} proposed the first graph-level black-box MIA in graph classification tasks, leveraging the model's prediction probability vectors to construct attack features and training a binary classifier. They also introduced multiple types of probability-based metrics to assess membership privacy leakage. Yang et al. \cite{10} proposed combining one-hot class labels with prediction probability vectors as attack features to train the attack model. And they utilized a class-dependent threshold to implement the attack based on modified prediction entropy. However, existing MIAs rely on the model providing prediction probability vectors and cannot be implemented when the model only outputs prediction labels.

Research on label-only membership inference attacks on graphs is still insufficient, but in non-graph domains, several studies have explored label-only MIAs \cite{17,18,21,22}. In the graph domain, Conti et al. \cite{11} utilized the fixed properties of the node, prediction correctness of the target model, node’s feature and neighborhood information to construct attack features, training an attack model to identify the membership of a node. However, graph-level label-only MIAs remain insufficiently explored. Therefore, we propose GLO-MIA, the first label-only membership inference attack against graph classification tasks.


\section{Graph-Level Label-Only Membership Inference Attack }\label{s4}
In this section, we introduce the threat model and implementation details of our GLO-MIA.
\subsection{Threat Model }
\subsubsection{Formalization of The Problem }
In the graph classification tasks of GNNs, the membership inference attacks aim to determine whether a given graph \( g_t \) belongs to the training set of the target GNN model. Formally, given a trained target GNN model \( f_t \) that only provides the prediction labels, a target graph \( g_t \), and the external knowledge \( \Omega \) obtained by the adversary, our attack \( \mathcal{A} \) can be expressed as:
\begin{linenomath}
\begin{equation}
\mathcal{A}: g_t, f_t, \Omega \to \{1, 0\}
\end{equation}
\end{linenomath}
where 1 indicates \( g_t \) is in the training set of \( f_t \), and 0 otherwise. 
\subsubsection{Adversary’s Knowledge and Capability}
\textbf{Label-only black-box setting.} We assume that the adversary cannot access the parameters or internal representations of the target model \(f_{t}\) and can only query the target model. The target model only provides prediction labels rather than prediction probability vectors. This represents a realistic and the most restrictive scenario, making it significantly more challenging for the adversary to attack. 

\textbf{Shadow Dataset and Shadow Model. }We assume the adversary owns a shadow dataset from the same domain as the target model's training dataset, with no overlap between the shadow and target datasets. Additionally, the adversary knows the architecture and training algorithm of the target model. The adversary can use the shadow dataset to train a shadow model with the same architecture as the target model to mimics its behavior. Notably, the membership and non-membership status of the shadow dataset in the shadow model is transparent to the adversary.
\subsection{Attack Methodology}
\subsubsection{Attack Intuition} \label{s421}
In the label-only setting, the adversary can only obtain the label-only output by the target model, making it impossible to rely on prediction probability vectors for the attack. Choquette-Choo et al. \cite{18} evaluated the robustness of the model’s prediction labels under perturbations of the input, to infer membership and their proposed MIAs could be applied in image, audio, and natural language domains. Inspired by their work, we infer membership of target graph by calculating the robustness score, which is defined as follows:
\begin{linenomath}
\begin{equation}
score_t = 
\begin{cases} 
\frac{\sum_{g_t' \in G_t} \mathbb{I}(f_t(g_t'), y_t)}{|G_t|}, & y_t = y \\ 
0, & \text{otherwise} \label{score} 
\end{cases}
\end{equation}
\end{linenomath}

where \( g_t \) represents a given target graph with the ground-truth label \( y \), \( score_t \) denotes the robustness score of \( g_t \), \( y_t \) is the predicted label of \( g_t \) by the target model \( f_t \), \( G_t \) is a set of perturbed graphs of \( g_t \), and \( g_t' \in G_t \) is one of perturbed graphs. \(|G_t|\) represents the number of elements in \( G_t \). The \(\mathbb{I}(\cdot)\) is the indicator function, specifically, if \( f_t(g_t') = y \), i.e., \( y_t = y \), it is recorded as 1, otherwise, it is 0. Graphs with relatively higher robustness scores are considered members.
\subsubsection{Attack Overview }
\begin{figure}
    \includegraphics[width=\textwidth]{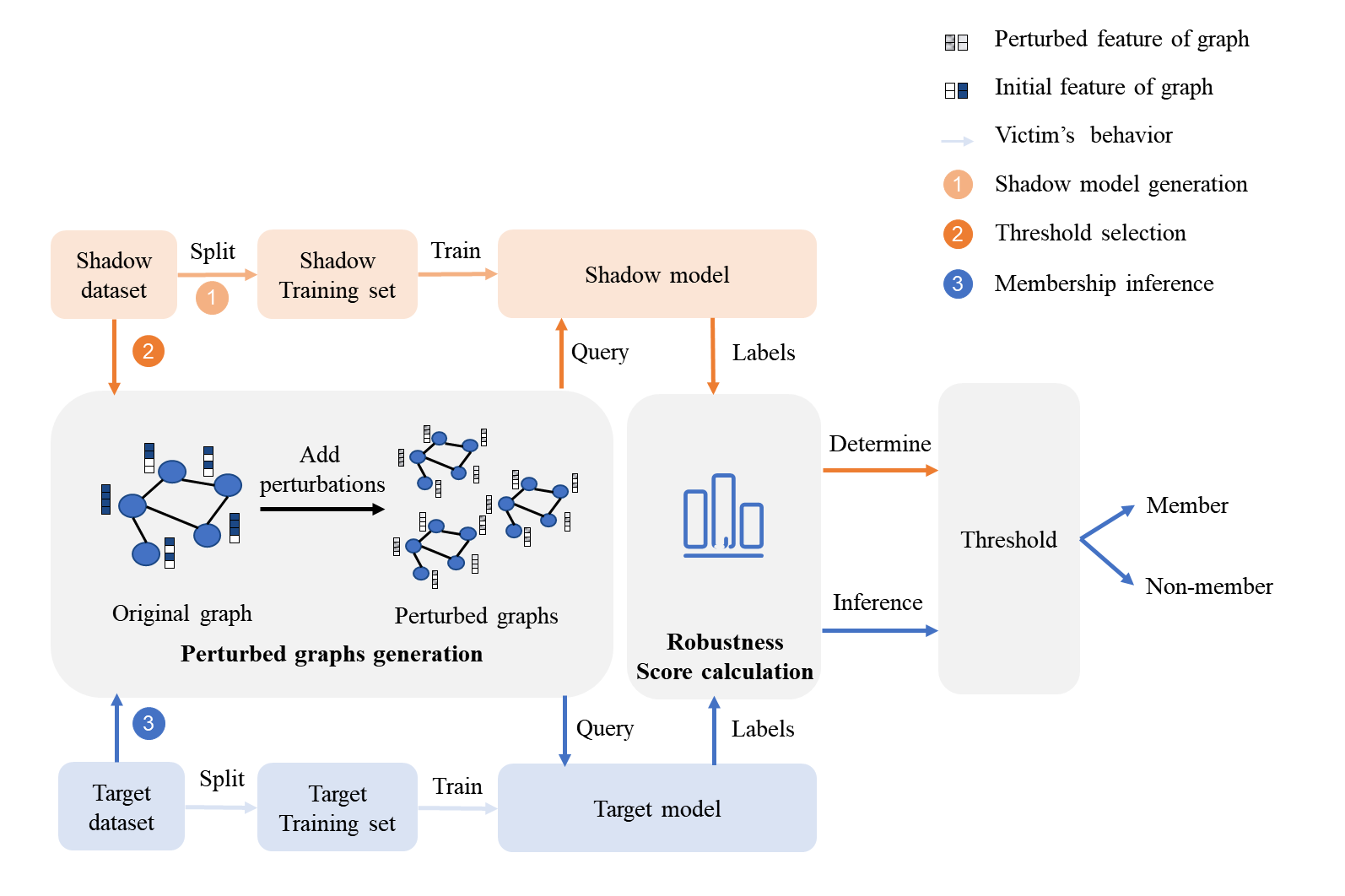}
    \caption{Illustration of GLO-MIA (step1: Shadow model generation, step2: Threshold selection, step3: Membership inference).} \label{fig1}
\end{figure}
Our attack process is shown in the Figure~\ref{fig1}, it consists of following four steps:

(1) \textbf{Shadow model generation}. The adversary splits the shadow dataset into training and testing sets and trains a shadow model on the training set to mimic the target model's behavior.  

(2) \textbf{Threshold selection}. 
    \begin{list}{}{\setlength{\leftmargin}{18mm}}
         \item [(a)] The adversary generates a set of perturbed graphs for each shadow graph using Algorithm ~\ref{alg1}, as introduced in Section~\ref{s423}.
         \item [(b)] The adversary calculates robustness scores using the prediction labels via Equation~\ref{score}, as introduced in Section ~\ref{s421} .
        \item [(c)] The adversary labels the robustness scores of the shadow training data as "member" and those of the shadow testing data as "non-member" to construct the attack dataset, by which a threshold of the robustness score can be selected.
    \end{list}

(3) \textbf{Membership inference}. 
    \begin{list}{}{\setlength{\leftmargin}{18mm}}
        \item[(a)] The adversary perturbs the target graph to generate a set of perturbed graphs.
        \item[(b)] The adversary calculates robustness score of the target graph.
        \item[(c)] The adversary campares the robustness sore of the target graph with the selected threshold. If the score is above the threshold, it's a member, otherwise, it's a non-member.
    \end{list}

\subsubsection{Perturbed Graphs Generation}\label{s423}
\textbf{Perturbed graphs generation}. We propose a perturbed graphs generation method to distinguish between members and non-members by perturbing the effective (i.e., non-zero) features on the graph, as described in Algorithm ~\ref{alg1}. Specifically, our method consists of the following steps:

(1) \textbf{Identify non-zero features}. Given a target graph \( g_{t} \), we identify its non-zero features. The positions of non-zero features  are indicated by \(MM\), which is a boolean matrix where the element value \texttt{true} indicates the presence of a non-zero value, while the element value \texttt{false} denotes a zero value (line 3 in Algorithm ~\ref{alg1}).

(2) \textbf{Adjust perturbation range}. The perturbation range is initialized from \(r_{min} = 0.1\) to \(r_{max} = 0.5\). A scaler \(s\) is used to adjust the perturbation range for different datasets. The perturbation range after adjustment is from \(s \times r_{min}\) to \(s \times r_{max}\) (line 4).  

(3) \textbf{Generate perturbed graphs}.
The process of generating perturbed graphs is as follows:  
    \begin{list}{}{\setlength{\leftmargin}{18mm}}
         \item[(a)]Create a copy of the input graph \(g_{t}\) as \(g_{t}'\) to ensure the perturbations are applied on the \(g_{t}'\) without affecting the \(g_{t}\).
         \item[(b)] Initialize perturbation matrix (denoted by \(PM\)) as a zero matrix of shape \(n\times d\), where \(n\) and  \(d\) represents the number of nodes and feature dimension of the input graph, respectively. It stores perturbation values, which are calculated as follows (line 7):
            \begin{equation}\label{pm}
            \mathrm PM[i][j] = 
            \begin{cases}
            p_{ij}, & \text{if } {MM}[i][j] = \mathrm{true} \\
            0, & \text{otherwise}
            \end{cases}, \quad \text{where } p \sim \mathcal{U}(r_{\min}, r_{\max})
            \end{equation}
            where $p_{ij}$ is a random perturbation value uniformly sampled from the perturbation range which is adjusted by scaler  \(s\).  
        \item[(c)] Initialize operator matrix (denoted by \(OM\)) as a zero matrix of shape \(n\times d\), and store the arithmetic operators, at the non-zero feature positions indicated by \(MM\), generate the random arithmetic operators according to the following equation (line 8) :
            \begin{equation}\label{om}
            \mathrm OM[i][j] = 
            \begin{cases}
            2 \cdot o_{ij} - 1, & \text{if } {MM}[i][j] = \mathrm{true} \\
            0, & \text{otherwise}
            \end{cases}
            \end{equation}
            where $o_{ij}$ is an integer randomly sampled between \(0\)  and \(1\). The values of elements of \(OM\) matrix can only be \(+1\) or \(-1\), representing addition or subtraction of a perturbation value.
         
         \item[(d)] Element-wise multiply perturbation values by their corresponding operators and add them to the features of \(g_{t}'\), ensuring that perturbations are added only to non-zero features, while others remain unchanged (lines 9-10).
         \item[(e)] Finally, add \(g_{t}'\) to the set \(G_{t}\) which is the set of perturbed graphs (line 11).
     \end{list}
   
   Repeat steps (a) to (e) until \(N\) perturbed graphs are generated.

The perturbations that we perform are irrelevant to the feature types, and our purpose is to distinguish between members and non-members rather than to cause severe model misclassification or to make the perturbations imperceptible.

\begin{algorithm}
\caption{Perturbed\_Graphs\_Generation ($g_t, N, s, r_{min}, r_{max}$)}
\label{alg1}
\begin{algorithmic}[1]
\Statex \textbf{Input:} 
    \texttt{$g_t$} --input graph,
    \Statex \hspace{3em}\texttt{$N$} -- the number of perturbed graphs,
    \Statex \hspace{3em}\texttt{$s$} -- scaler,
    \Statex \hspace{3em}\texttt{$r_{min}, r_{max}$} -- perturbation range
\Statex \textbf{Output:} \texttt{$G_t$} -- set of generated perturbed graphs
\end{algorithmic}
\begin{algorithmic}[1]
\State $n, d, X \gets g_t$ \Comment{Number of nodes, feature dimensions, and node feature matrix}
\State $G_t \gets \emptyset$

\State $MM \gets X \neq 0$ \Comment{Generate non-zero mask} 
\State $ r_{\min}, r_{\max}\gets s \times r_{\min}, s \times r_{\max}$ \Comment{Adjust perturbation range} 
\For{$i \in \{1, \dots, N\}$} \label{line:for_loop}
    \State $g_t' \gets g_t$
    \State $PM \gets \texttt{Generate perturbation values via Eq.\ref{pm} with adjusted $r_{\min}$ and $r_{\max}$}$
    \State $OM \gets \texttt{Generate arithmetic operators via Eq.\ref{om}} $
    \State $PM \gets PM \odot OM$ \Comment{Generate perturbation} 
    \State $X_{g_t'} \gets PM + X_{g_t'}$ \Comment{Apply perturbation on node features of \(g_{t}'\)}
    \State $G_t \gets G_t \cup \{g_t'\}$
\EndFor
\State \textbf{return} $G_t$
\end{algorithmic}
\end{algorithm}
\subsubsection{Perturbation Magnitude Estimation and Threshold Selection}\label{s424}
As discribed in Algorithm ~\ref{alg1}, a scaler \(s\) is used to adjust the perturbation range which controls the perturbation magnitude of graphs. However, perturbations that are either too large or too small fail to identify the membership of the target graph, which is validated in Section \ref{s53}. Therefore, we need to find appropriate perturbations to distinguish between members and non-members, which can be roughly estimated using shadow dataset and model.

During estimation, \(s\) is gradually increased in small steps. At each step:

(1) Calculate the robustness scores of the shadow data using the current \(s\).

(2) Label the robustness scores of the shadow training data as "member",  while labeling those of the shadow testing data as "non-member" to construct the attack dataset.

(3) Maximize the classification accuracy of distinguishing between shadow members and non-members under current \(s\) by iteratively selecting threshold using the ROC curve. Record the current AUC and accuracy which jointly reflect the classification performance of the estimation. 

(4) If both the  AUC and accuracy begin to decline, the process terminates, and the last \(s\) and threshold before the decline are recorded for use in the attack on the target model. Otherwise, increment \(s\) by a small step and repeat steps (1) to (3).

\section{Experiments}\label{s5}
In this section, we systematically evaluate the performance of GLO-MIA. In Section ~\ref{s51}, we introduce the experimental setup. In Section ~\ref{s52}, we evaluate GLO-MIA using three graph classification datasets and four GNN models, and compare the experimental results with those of the baselines. In Section ~\ref{s53} and Section ~\ref{s54} , we analyze the impact of perturbation magnitude and the number of perturbed graphs on the performance of the proposed MIA, respectively.
\subsection{Experimental Setup}
\label{s51}
\subsubsection{Datasets }
TUDataset \cite{23} is a collection of benchmark datasets for graph classification and regression tasks. We evaluate GLO-MIA on three publicly available bioinformatics datasets from TUDataset, including DD \cite{24}, ENZYMES \cite{25,26}, and PROTEINS\_full \cite{24,25}. Each dataset consists of multiple independent graphs, which have different numbers of nodes and edges. Each graph is labeled for graph classification tasks. 

The DD dataset is used to classify the protein structures into enzymes and non-enzymes. The labels in PROTEINS\_full dataset show whether a protein is an enzyme. The ENZYMES assigns enzymes into six classes, which reflect the catalyzed chemical reaction. The statistical details of these datasets are provided in Table~\ref{tab1}.

We randomly divide the dataset into two non-overlapping subsets, including target dataset and the shadow dataset. To ensure fairness, we use training sets and testing sets of the same size.

\begin{table}[htbp]
\centering
\caption{Statistical information of datasets.}
\label{tab1}
\begin{tabular}{cccccc}
\toprule
\textbf{Dataset} & \textbf{Type} & \textbf{Graphs} & \textbf{Classes} & \textbf{Avg. Nodes} & \textbf{Avg. Edges} \\
\midrule
DD              & Bioinformatics & 1178            & 2                & 284.32              & 715.66              \\
PROTEINS\_full  & Bioinformatics & 1113            & 2                & 39.06               & 72.82               \\
ENZYMES         & Bioinformatics & 600             & 6                & 32.63               & 62.14               \\
\bottomrule
\end{tabular}
\end{table}

\subsubsection{Model Architectures and Training Settings}
The experimental models are four popular GNN models, including GCN \cite{3}, GAT \cite{12}, GraphSAGE \cite{13}, and GIN \cite{6}. All of these models are based on a 2-layer architecture, which is a common setting in practice. We use the cross-entropy loss function and the Adam optimizer for training.
\subsubsection{Metrics}
We adopt attack accuracy (ACC) and the area under the ROC curve (AUC) as metrics to evaluate the attack performance of GLO-MIA.

(1)\textbf{Attack accuracy (ACC):} Attack accuracy (ACC) represents the percentage of samples whose membership are inferred correctly among total samples by MIAs. It is as follows:
\begin{linenomath}
\begin{equation}
ACC = \frac{\text{the number of samples inferred correctly}}{\text{total samples}}
\end{equation}
\end{linenomath}

(2)\textbf{The area under the ROC curve (AUC):} The area under the ROC curve (AUC) is computed as the area under the curve of attack receiver operating characteristic (ROC), which is a threshold-independent metric.  

\subsubsection{Baselines}
Our evaluation aims to confirm whether GLO-MIA outperforms the gap attack \cite{14} and achieves performance comparable to probability-based MIAs based on prediction probability vectors \cite{9,10}, even the adversary only uses the prediction labels rather than prediction probability vectors.

(1) \textbf{Gap attack} \cite{14}. This is a naïve baseline commonly used for label-only MIAs \cite{17,18}. It determines membership based on whether the target model's prediction label is correct. Samples that are correctly predicted are regarded as members, while those incorrectly predicted are considered non-members. The attack accuracy of the gap attack is directly related to the gap between training and testing accuracy of the target model. If the adversary's target data are equally likely to be members or non-members of the training set, the attack accuracy of the gap attack can be calculated as follows \cite{18}:
\begin{linenomath}
\begin{equation}
ACC_{gap} = \frac{1}{2} + \frac{acc_{train} - acc_{test}}{2}
\end{equation}
\end{linenomath}

where \(acc_{train}\) and \(acc_{test}\) are the accuracy of the target model on the training dataset and the testing dataset, respectively. If the model overfits on its training data, which means it has a higher accuracy on the training set, the gap attack will achieve non-trivial performance.

(2) \textbf{Probability-based MIAs}. We implement two advanced probability-based MIAs which utilize the prediction probability vectors of the target model in graph classification tasks.
    \begin{list}{}{\setlength{\leftmargin}{18mm}}
        \item [(a)] \textbf{Cross-entropy-based MIA} \cite{9}. This method assumes that the loss of members should be smaller than that of non-members.
        \item [(b)] \textbf{Modified prediction entropy-based MIA} \cite{10}. This method assumes that members are expected to have lower modified prediction entropy than non-members.
    \end{list}

\subsection{Experimental Results}
\label{s52}
We train a shadow model with the same architecture as the target model using a non-overlapping shadow dataset that has the same distribution as the target dataset. Then we apply GLO-MIA on different GNN models. We generate 1000 perturbed graphs for each graph, to avoid the impact of randomness, we conduct our attack 5 times in all experiments and report the average results. 

We compare GLO-MIA with the gap attack, and the results are shown in Table ~\ref{tab2}. In Table ~\ref{tab2}, we record “train-test gap”, which represents the gap between the training and testing classification accuracy to measure the overfitting of target model. The best attack performance of each dataset is highlighted in bold. The experimental results show that GLO-MIA is effective on different GNN models and datasets. For example, the GraphSAGE trained on the DD dataset can achieve attack accuracy of 0.738, which is 5.6\% higher than the gap attack. On the ENZYMES dataset, the highest attack accuracy can reach 0.825, which is 8.5\% higher than the gap attack. Furthermore, our experimental results indicate that the attack performance on GIN models is comparatively lower than that on other GNN models, suggesting that GIN may have higher inherent robustness against our attack. In contrast, GraphSAGE is more vulnerable to our attack, likely due to its aggregation mechanism being more sensitive to feature perturbations on graphs.

\begin{figure}[H]
\centering
\begin{adjustwidth}{-\extralength}{0cm}
\subfloat[\centering DD]{\includegraphics[width=6cm]{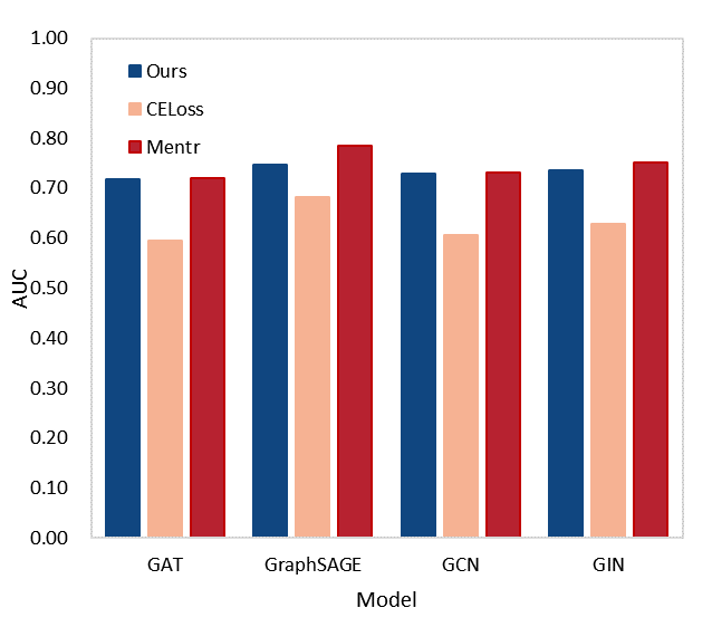}}
\hfill
\subfloat[\centering ENZYMES]{\includegraphics[width=6cm]{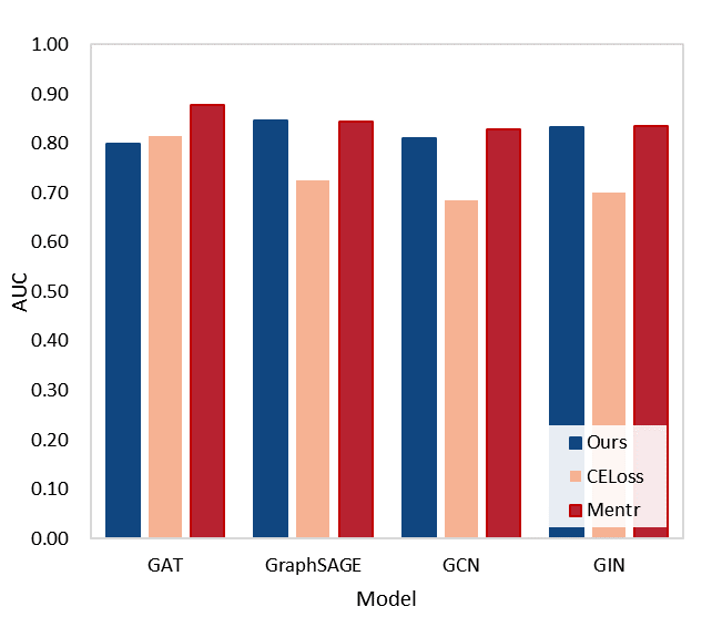}}
\hfill
\subfloat[\centering PROTEINS\_full]{\includegraphics[width=6cm]{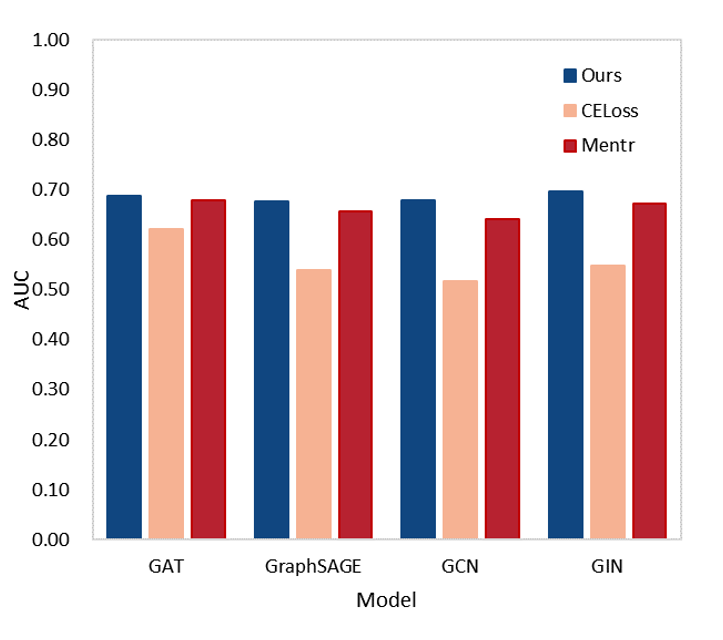}}
\end{adjustwidth}
\caption{Visualization of Table \ref{tab3}.\label{fig2}}
\end{figure}

Moreover, we compare GLO-MIA with two advanced threshold-based MIAs against graph classification tasks using prediction probability vectors under the same conditions as in Table ~\ref{tab2}. Since the threshold selection strategies are different and AUC is a threshold-independent evaluation metric, we adopt AUC to assess the performance of GLO-MIA and the probability-based MIAs. The performance of these MIAs is shown in Table ~\ref{tab3} and we further visualize the results in Figure ~\ref{fig2}. From the figure, we can see that our attack achieves close performance to that of the probability-based MIAs, and in most cases, outperforms cross-entropy-based MIA. This further confirms the effectiveness of GLO-MIA.
\begin{table}[H]
\centering
\caption{Comparison of attack accuracy between our attack and the gap attack.}
\label{tab2}
\begin{tabularx}{\textwidth}{CCCCC}
\toprule
\textbf{Dataset} & \textbf{Model} & \textbf{Train-test gap} & \textbf{Ours} & \textbf{Gap attack} \\
\midrule
DD              & GAT            & 0.360                  & 0.728         & 0.680               \\
                & GraphSAGE      & 0.364                  & \textbf{0.738} & 0.682               \\
                & GCN            & 0.352                  & 0.722         & 0.676               \\
                & GIN            & 0.376                  & 0.713         & 0.688               \\
ENZYMES         & GAT            & 0.480                  & 0.789         & 0.740               \\
                & GraphSAGE      & 0.480                  & \textbf{0.825} & 0.740               \\
                & GCN            & 0.520                  & 0.800         & 0.760               \\
                & GIN            & 0.547                  & 0.786         & 0.773               \\
PROTEINS\_full  & GAT            & 0.272                  & 0.676         & 0.636               \\
                & GraphSAGE      & 0.292                  & 0.670         & 0.646               \\
                & GCN            & 0.312                  & 0.674         & 0.656               \\
                & GIN            & 0.340                  & \textbf{0.697} & 0.670               \\
\bottomrule
\end{tabularx}
\end{table}
\begin{table}[htbp]
\centering
\caption{Comparison of AUC with two probability-based MIAs (CELoss: cross-entropy-based MIA, Mentr: modified prediction entropy-based MIA).}
\label{tab3}
\begin{tabularx}{\textwidth}{CCCCC}
\toprule
\textbf{Dataset} & \textbf{Model} & \textbf{Ours} & \textbf{CELoss} & \textbf{Mentr} \\
\midrule
DD              & GAT            & 0.717         & 0.595           & \textbf{0.721}           \\
                & GraphSAGE      & 0.746         & 0.682           & \textbf{0.785}           \\
                & GCN            & 0.728         & 0.606           & \textbf{0.731}           \\
                & GIN            & 0.736         & 0.629           & \textbf{0.751}           \\
ENZYMES         & GAT            & 0.798         & 0.816           & \textbf{0.877}           \\
                & GraphSAGE      & \textbf{0.845}         & 0.725           & 0.844           \\
                & GCN            & 0.809         & 0.685           & \textbf{0.829}           \\
                & GIN            & 0.832         & 0.701           & \textbf{0.834}           \\
PROTEINS\_full  & GAT            & \textbf{0.688}         & 0.621           & 0.678           \\
                & GraphSAGE      & \textbf{0.677}         & 0.540           & 0.656           \\
                & GCN            & \textbf{0.679}         & 0.516           & 0.641           \\
                & GIN            & \textbf{0.697}         & 0.548           & 0.672           \\
\bottomrule
\end{tabularx}
\end{table}

\subsection{The Impact of Perturbation Magnitude}
\label{s53}
\begin{figure}[H]

\centering
\subfloat[\centering DD]
{\includegraphics[width=0.5\textwidth]{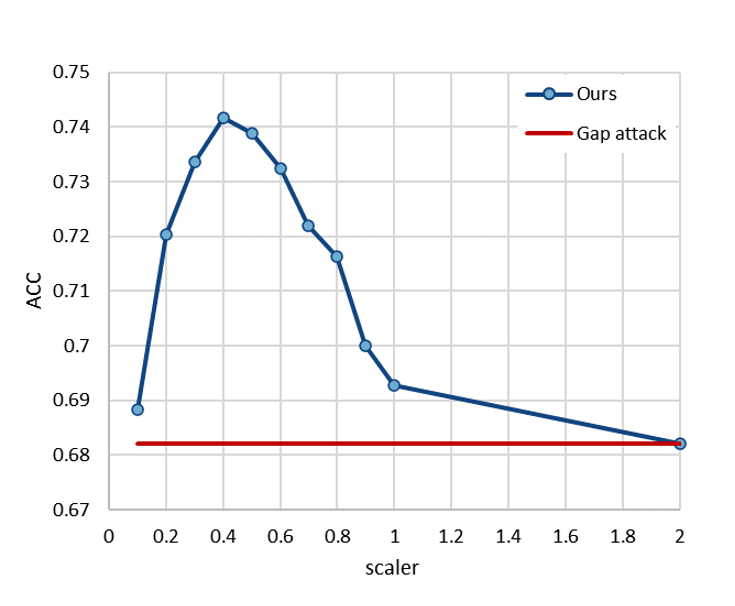}}
\hspace{-3pt}
\subfloat[\centering ENZYMES]
{\includegraphics[width=0.5\textwidth]{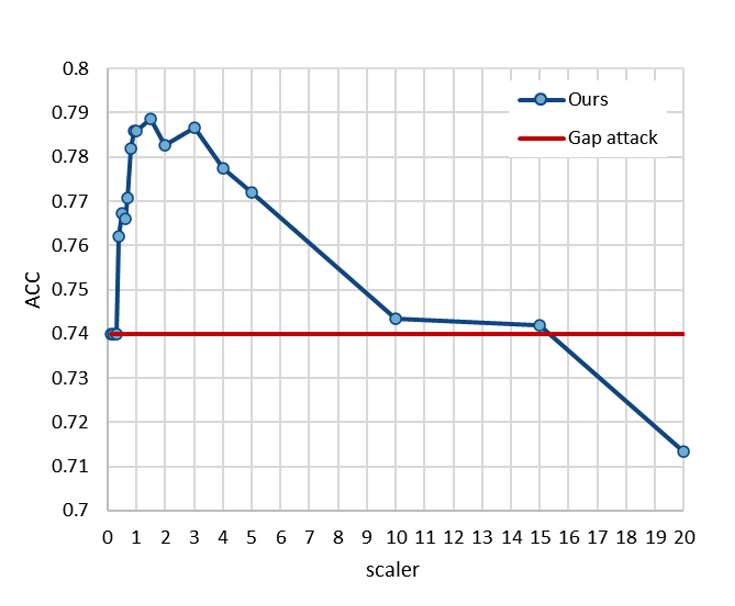}}
\caption{Impact of perturbation magnitude on the performance of the attacks. 
(red lines: gap attack, x-axis: the scaler size, y-axis: attack accuracy).}
\label{fig3}
\end{figure}

In this section, we analyze the impact of the perturbation magnitude on GLO-MIA. In our attack, the perturbation magnitude is controlled by adjusting the scaler size. We evaluate the performance of our attack under the different scaler sizes, using the ENZYMES and DD datasets. The results are shown in Figure ~\ref{fig3}, where the x-axis represents the scaler size and the y-axis represents attack accuracy. The red and blue lines represent the attack accuracy of the gap attack and our attack, respectively. From the figure, we observe that as the size of scaler increases, the attack accuracy initially improves and then declines.

\begin{figure}[H]
\centering
\begin{adjustwidth}{-\extralength}{0cm}
\subfloat[\centering s=0.1]{\includegraphics[width=6cm]{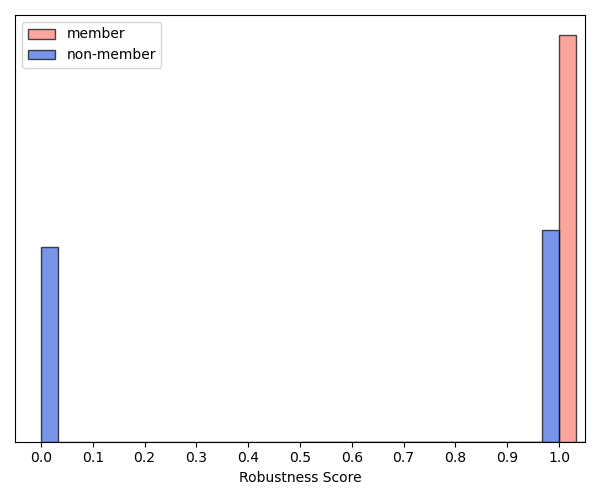}}
\hfill
\subfloat[\centering s=1.5]{\includegraphics[width=6cm]{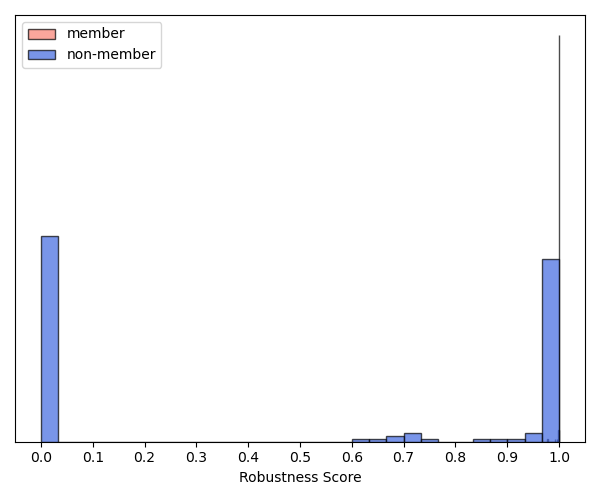}}
\hfill
\subfloat[\centering s=15.0]{\includegraphics[width=6cm]{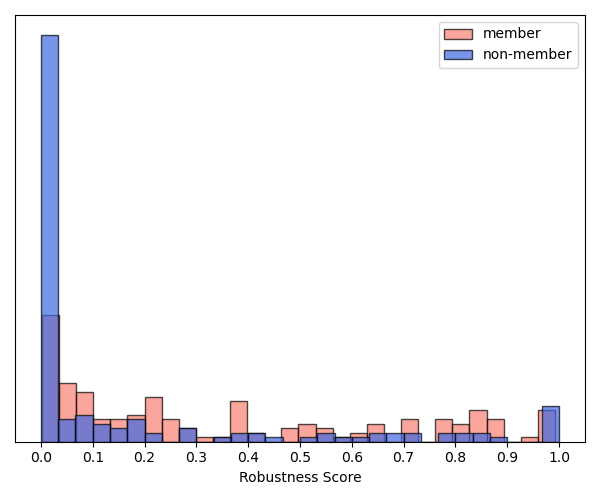}}
\end{adjustwidth}
\caption{Robustness score distributions of members and non-members under different perturbations (red: members, blue: non-members).\label{fig4}}
\end{figure}

To analyze this phenomenon, we further plot the robustness score distributions of members and non-members on the ENZYMES dataset with scaler sizes of 0.1, 1.5, and 15.0, as shown in Figure ~\ref{fig4}. The red and blue bars represent the robustness score of member and non-member, respectively. We observe that when the perturbation is too small (s=0.1), the model predictions are not effectively disturbed, resulting in highly overlapping robustness score distributions of members and non-members, and the attack accuracy remains similar to the gap attack. When the perturbation is effective (s=1.5), the robustness scores of non-members decrease significantly, while members maintain relatively high robustness, leading to clear distinction between the two distributions. At this point, our attack accuracy outperforms the baseline by approximately 5\%. However, when the perturbation is too large (s=15.0), both members and non-members are misclassified, causing their distributions to overlap again and significantly reducing the attack performance, even lower than the gap attack (s=20.0). 

The experimental results demonstrate that the choice of perturbation magnitude is a critical factor for the performance of our MIA attack, and suitable perturbations can effectively enhance the distinguishability between members and non-members.

\subsection{The Impact of The Number of Perturbed Graphs}
\label{s54}
\begin{figure}
\centering
\subfloat[\centering DD]
{\includegraphics[width=0.5\textwidth]{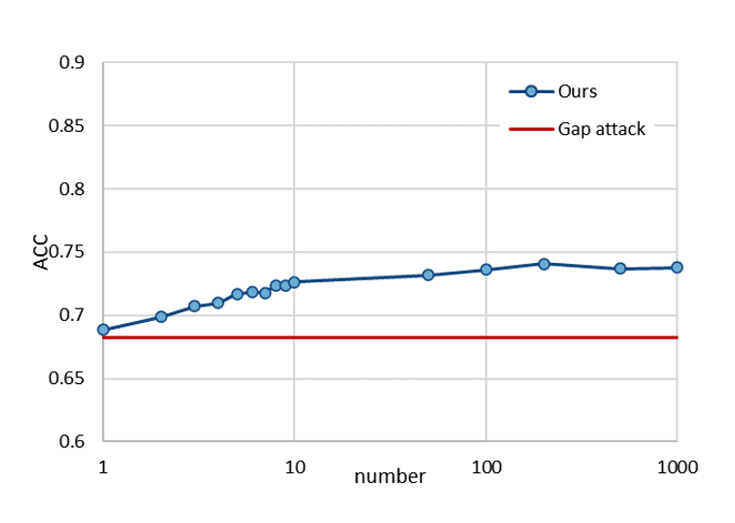}}
\hspace{-3pt}
\subfloat[\centering ENZYMES]
{\includegraphics[width=0.5\textwidth]{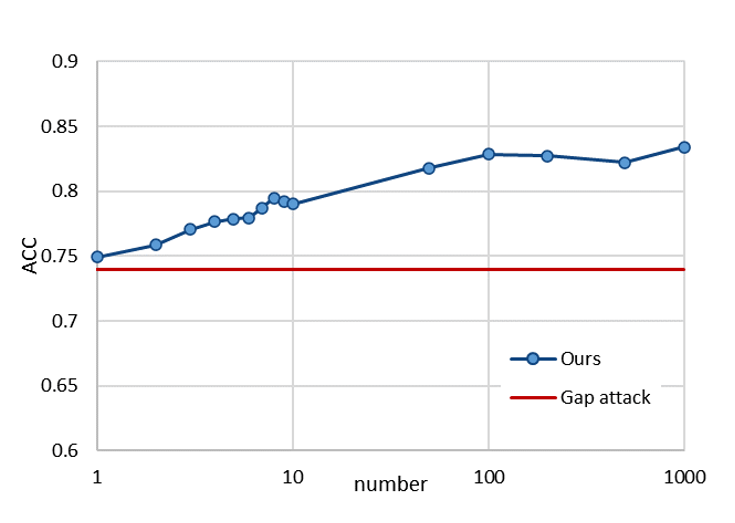}}
\caption{Impact of the number of perturbed graphs on the performance of the attacks. (red line: gap attack, x-axis: the number of perturbed graphs, y-axis: attack accuracy).}
\label{fig5}
\end{figure}

We further investigate the impact of the number of perturbed graphs on our attack performance, evaluating it using the GraphSAGE model trained with the ENZYMES and DD datasets and recording the membership inference attack accuracy. The results are shown in Figure ~\ref{fig5}, where the x-axis represents the number of perturbed graphs and the y-axis represents attack accuracy, the red and blue lines represent the accuracy of the gap attack and our attack, respectively. From the figure, we observed that as the number of perturbed graphs increases, the attack performance generally improves, indicating that increasing the number of perturbed graphs enhances the attack’s effectiveness. When the number of perturbed graphs is small (e.g., 1 to 10), the attack accuracy is relatively low but still effective, which means that GLO-MIA remains a threat even with limited query budgets. In practical applications, if computational resources permit, increasing the number of perturbed graphs can improve the attack accuracy.

\section{Conclusion and Future Work}\label{s6}
This paper proposes the first label-only MIA called GLO-MIA against GNNs in graph classification tasks, revealing the vulnerability of GNNs to such threats. Based on the intuition that the target model's predictions on the training data are more stable than those on testing data, GLO-MIA can effectively infer the membership of given graph samples by comparing the their robustness scores with a predefined threshold. Our empirical evaluation results on four popular GNN models (GCN, GAT, GraphSAGE and GIN) and three benchmark datasets show that the attack accuracy of GLO-MIA can reach up to 0.825 at best, with a maximum improvement of 8.5\% compared with the gap attack, and the performance of GLO-MIA is close to that of probability-based MIAs. Furthermore, we analyze the impact of perturbation magnitude and the number of perturbed graphs on the attack performance. In the future, we will study other label-only MIAs in graph classification tasks and the defense methods to these MIAs.

\begin{adjustwidth}{-\extralength}{0cm}
\reftitle{References}

\isAPAandChicago{}{%

}
\end{adjustwidth}
\end{document}